# Multimodal image registration for effective thermographic fever screening

C.Y.N. Dwith[a,b], Pejhman Ghassemi[a], Joshua Pfefer[a], Jon Casamento[a] and Quanzeng Wang*[a]

[a] U.S. Food and Drug Administration, Center for Devices and Radiological Health, 10903 New Hampshire Avenue, Silver Spring, Maryland 20993, USA

[b] University of Maryland, College Park, Maryland 20740, USA

*Quanzeng.Wang@fda.hhs.gov

**ABSTRACT**

Fever screening based on infrared thermographs (IRTs) is a viable mass screening approach during infectious disease pandemics, such as Ebola and SARS, for temperature monitoring in public places like hospitals and airports. IRTs have found to be powerful, quick and non-invasive methods to detect elevated temperatures. Moreover, regions medially adjacent to the inner canthi (called the canthi regions in this paper) are preferred sites for fever screening. Accurate localization of the canthi regions can be achieved through multi-modal registration of infrared (IR) and white-light images. We proposed a registration method through a coarse-fine registration strategy using different registration models based on landmarks and edge detection on eye contours. We evaluated the registration accuracy to be within ± 2.7 mm, which enables accurate localization of the canthi regions.

## 1. INTRODUCTION

Mitigation of the threat of infectious pandemics such as Ebola virus disease may be possible through mass screening in public places such as airports. Fever screening based on non-contact infrared thermometers (NCITs) and IRTs represents the only currently viable mass screening approach. The IEC 80601-2-59 international standard [1] indicates that regions medially adjacent to the inner canthi (called the canthi regions in this paper) provide accurate estimates of core body temperature and are preferred sites for fever screening. Therefore, rapid, automated identification of the canthi within facial IR images (also called thermal images) may greatly facilitate fever screening. A common method for detecting canthi in an IR image is through multi-modal image registration (MMIR) [2, 3] of simultaneously captured IR and white-light images (also called visible images). IR images provide temperature information while white-light images are rich with exploitable features for automatic canthi detection In this paper, we focused on registration of IR and white-light face images for fever screening. We implement an image registration method, which enables low cost integration of available IR and white-light images from cameras without any additional requirements for hardware.

### 1.1 IR Thermography

IR thermography detects IR radiation and produces images of that radiation, called thermograms. The object surface temperature governs the amount of IR radiation emitted from the surface. The IEC 80601-2-59 international standard indicates that the canthi regions (Fig.1) provide accurate estimates of core body temperature, since the canthi regions are supplied by the internal carotid artery and not susceptible to external environmental changes. Therefore, rapid, automated identification of the canthi regions within facial IR images may greatly facilitate fever screening.

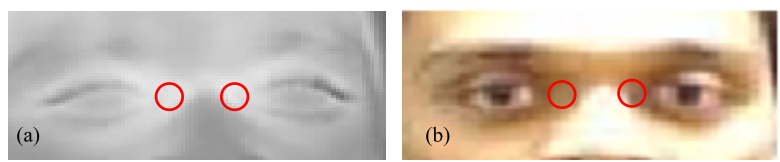

Fig.1. The canthi regions for fever screening (a: an IR image; b: a white-light image)

## 1.2 Image Registration

Image registration [2, 3], especially MMIR, is important for medical imaging such as magnetic resonance imaging (MRI), computed tomography (CT), positron emission tomography (PED), and IR imaging. MMIR is an effective way of combing images that provide different information into one combined spatial reference frame. Image registration algorithms are mainly composed of three components: i) similarity metric, ii) transformation model and iii) optimization algorithms. *Similarity metric* is used in image registration to match objects in related images taken from different times, modalities, viewpoints, scales and resolutions. Most registration methods use global statistical measure to align the images.[2] Appearance features such as gray levels, colors, and textures are preferred for similarity matching. In case of white-light and IR images which is manifested by different phenomena, these features will not likely match. However, certain features salient structures could be adopted, such as points of high curvature, line intersections, strong edges, structural contours and silhouettes within the images. In this paper, we used edge map, a significant common feature in the face images, as the similarity metric. *Transformation models* for image registration can be broadly classified into rigid and non-rigid transformation models. The combination of translation, rotation and scaling together defines global rigid transform. Non-rigid parametric models are used to describe transformations that define free-form deformations using vector fields [2, 3] . One of the significant contributors to transformation performance is the interpolation technique. The estimated transformation on an image will result in non-integer grid locations, which needs interpolation techniques to compute the moving images intensity values at the specified locations. Even though a high order interpolator performs better, we choose bicubic interpolation as tradeoff between performance and computation.[2] A detailed discussion on optimization techniques is beyond the scope of this article. In this study, we use regular step gradient decent based optimization [2, 3], which is quite widely used iterative optimization problem.

## 2. METHODS

Human face has non-rigid shape and the deformation of the point set sampled from face is restricted by physical constraints of bones and muscles. Therefore, parametric models such as rigid transformation [2, 3] cannot produce accurate alignment. To address this problem, we approach a two-step registration strategy with coarse and fine registrations. Coarse registration is used for alignment of images and detection of region of interest (ROI). Fine registration is used to improve the registration accuracy to enable accurate detection of the canthi regions from an IR image. The proposed method tries to implement automated temperature measurement using canthi regions. This needs to have accurate IR-visible face registration for localization of canthi regions. The scheme for proposed algorithms is presented in Fig.2. Implementation of image registration is divided into two steps: i) Coarse Registration, and ii) Fine Registration.

## 2.1 Coarse registration

Coarse registration is used for selecting ROI around the face and is primarily based on mutual information (MI). MI as metric for image registration was first proposed by Woods *et al.* [4, 5]. In case of multi-modal registration, regions with different gray levels (*i.e.* ratio of gray values) in an image would correspond to similar regions in another image that also contain similar number of grey values (maybe of different value). Ideally, the correspondence between these gray levels might not change significantly across either of these images. Hill *et al.* [6] proposed registration by constructing a joint histogram, which is defined as two-dimensional plot showing combinations of grey values as shown in Fig.3. At location (i, j) in joint histogram, gives number of intensity values we have encountered that have intensity i in the first image and intensity j in second image. Joint histogram shows increased dispersion as mis-registration of images increases. Shannon entropy (referred as entropy) is used as registration metric to measure dispersion in joint histogram. Entropy of discrete random variable $X_i$ with probability distribution function $P(X_i)$ is given by:

$$H(X) = \sum_{i=1}^{n} P(X_i) \log(\frac{1}{P(X_i)})$$

Entropy does not depend on value of random variable, but depends on distribution of random variable. This definition of entropy can be extended to images, where the probability distribution function is constructed using histogram distribution of pixel values in the image. Entropy of joint histogram decreases as the alignment of images increases as shown in Fig.3. We now define MI as I(A, B) based on the entropy of two input images A and B, which relates to joint histogram entropy as follows [3]:

$$I(A,B) = H(A) + H(B) - H(A,B)$$

where H(A), H(B) and H (A, B) are entropy of image A, entropy of image B and joint histogram entropy respectively. The above equation related minimization of joint histogram to maximization of MI. Hence, problem of registration is converted into optimization problem that tries to maximize the MI using different transformations.

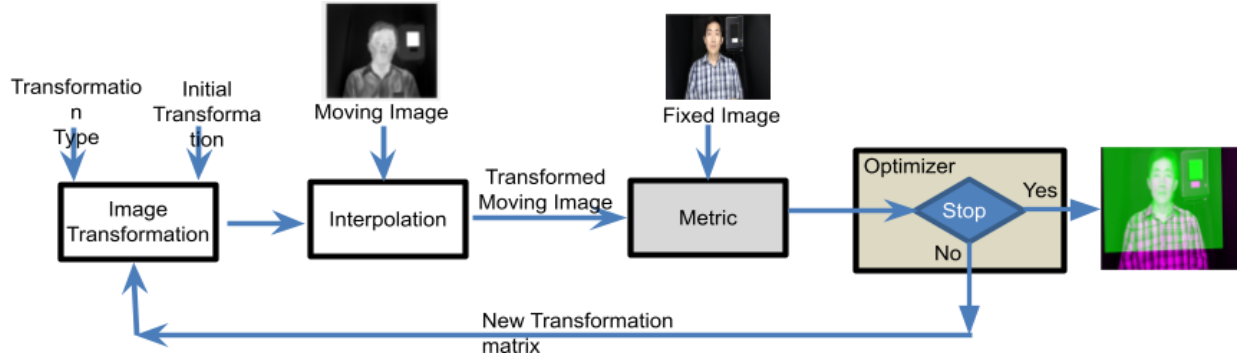

Fig.2. Block diagram of the system for image registration based on Mutual Information

Fig.3. Joint Histogram Plot (a: Not Registration Images; b: Registered Images (Ideal))

Coarse registration algorithm is implemented in Matlab using Mattes MI algorithm [7]. In this algorithm, single intensity pixel/sample is drawn from image. The marginal and joint probability density function (PDF) evaluated at discrete position using samples of pixel intensities. Optimizer used for implementation of this algorithm is the regular step gradient descent optimizer [8].

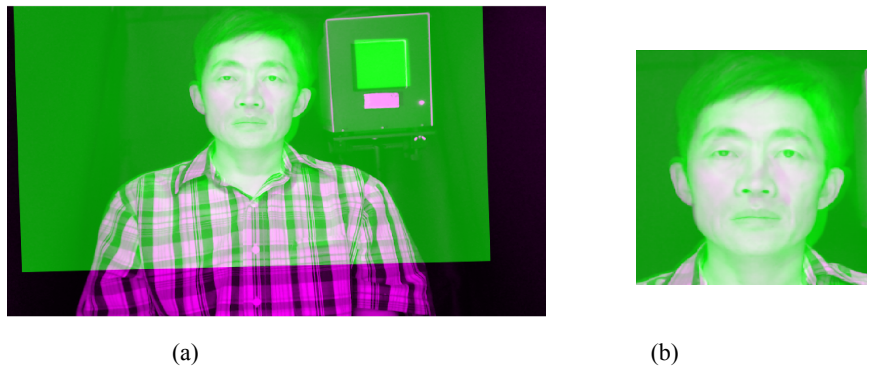

(a)            (b)

Fig.4. Coarse Image Registration (a: whole images; b: ROI)

## 2.2 Fine Registration

We uses fine registration to improve registration accuracy of the IR-visible images and enable accurate localization of the canthi regions. Unlike the rigid transforms [2, 3] used for coarse registration, the fine registration uses free-form deformations and needs to define a vector field for each pixel in the image. The fine registration is implemented through the Demons Algorithm [9, 10], a free-form deformation widely used in medical imaging.

Thirion [9] proposed non-rigid registration as a diffusion process, which introduces entities called Demons that exert forces according to local characteristics of images. These forces were inspired from optical flow equations [11]. In this paper, we discuss non-rigid registration as an iterative optimization problem. The basic idea of the Demons algorithm for non-rigid registration is that static reference image act as local forces that move the pixels in moving image to match static image. During each iteration, the moving image is transformed using the moving vector *dV = (dx, dy)* for each pixel as shown [10]:

$$dV^{(n+1)} = \frac{(I_{mov}^{(n)} - I_{Ref}^{(0)}) \times \nabla I_{Ref}^{(0)}}{(I_{mov}^{(n)} - I_{Ref}^{(0)}) + |\nabla I_{Ref}^{(0)}|}$$

where $I_{Ref}^{(n)}$ and $I_{mov}^{(n)}$ are intensity of static and images at the n$^{th}$ iteration. $I_{Ref}^{(0)}$ and $I_{mov}^{(0)}$ represent the original static and moving images, respectively. Gaussian filter is used to smooth the displacement fields, which enables noise suppression and preserves the geometric continuity of deformed image. The gradient of static image $\nabla I_{Ref}^{(0)}$ is computed only once during the iterations. Moreover, the Demons algorithm assumes that displacement vector is reasonably small or local. However, in real clinical cases, such assumption is violated. To reduce the magnitude of displacement vectors, we use multi-scale approach, where both static and moving images are down-sampled to low resolution images. Similarly, the displacement fields at each stage are up-sampled from finer scale. This also enables a large computational advantage for large image sizes.

However, the Demons algorithm is applicable only to mono-modal image registration [2, 3]. In this paper, we generate edge maps for IR and white-light images using canny edge detector for non-rigid transformation. These edge maps emphasize the contour edges of face and show good similarity between thermal and white-light images [12]. The eye regions are used to predict the free-form transformation used for non-rigid registration. As an iterative optimization method, we need to define effective stopping criteria. In this study, we used a tolerance criteria that stops the iterations if the mean squared error (MSE) is increasing with the iterations and also if the decrease in threshold for each iteration is within a convergence tolerance.

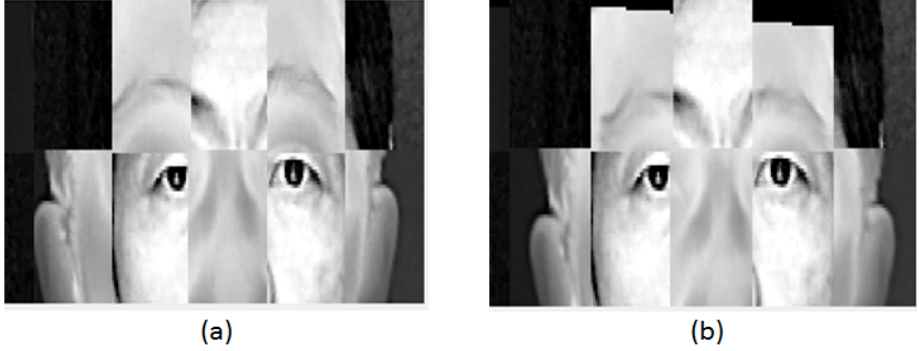

Fig.5. Registered images using checkered view (a: Coarse Registration; b: Fine Registration)

## 3. RESULTS

The images (Fig.6) of volunteers were captured using three different cameras: i) an IR camera from FLIR Systems Inc. (IRT1), ii) an IR camera from Infrared Cameras Inc. (IRT2), and iii) a white-light camera from Logitech. These images are used as input images for image registration.

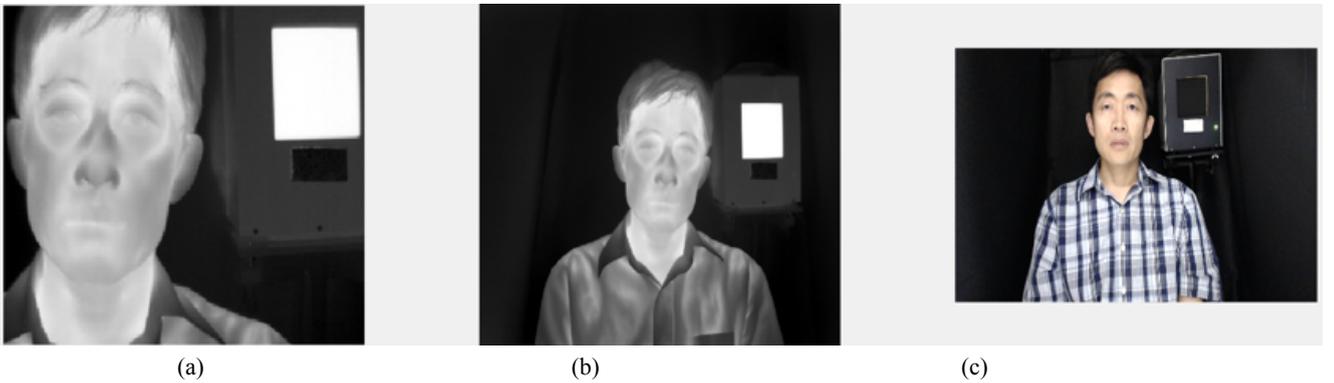

Fig.6. Images from three cameras (a: IRT1; b: IRT2; c: white-light camera)

We quantitatively evaluated the coarse-fine registration strategy of IR and white-light images. Circular aluminum foils with diameter of 7 mm were attached to different locations around the canthi regions of the volunteer as landmarks. The captures IR and white-light images were used as the input images for image registration. The landmarks and their correspondence between image pairs were manually selected from the input images as control points. After image registration, the mean square error (MSE) of distances between each pair of control points in the IR and white-light images was calculated as a qualitative performance metric. The average matching error of the coarse registration is around 12.96, 11.39 and 14.48 pixels on three individuals. It improves significantly to average error of about 4.19, 4.27 and 6.97 pixels for fine registration. The scale factor of image is measured to be ~1.2 mm/pixel. Absolute mean error in localization of canthi region, after taking square root and multiplication with scale factor of 1.2mm/pixel, is +/- 2.7mm in localization of canthi region. This shows promising results to enables automated and accurate localization of canthi region for temperature measurement in fever screening.

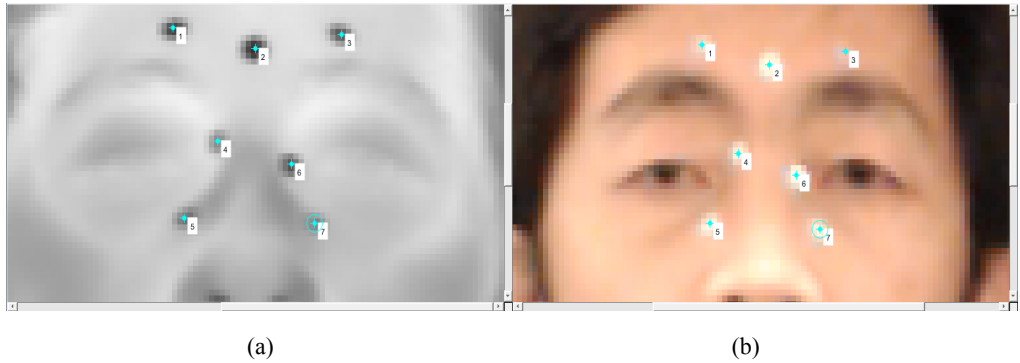

(a) (b)

Fig.7. Control point selection for registration evaluation (a: IR image; b: white-light image)

Table 1. Using control points in white-light and IR Image for registration accuracy

| Subject No. | Coarse Registration | Fine Registration |
|---|---|---|
| Subject 1 | 12.96 | 4.19 |
| Subject 2 | 11.39 | 4.27 |
| Subject 3 | 14.48 | 6.97 |

## CONCLUSION

In this paper, we propose a system that uses rigid and non-rigid registration algorithms for IR and white-light face images for detection and temperature measurement using the canthi regions. We study the accuracy of non-rigid registration that influences the localization of the canthi regions. The results suggest that the image registration method can handle localization of canthi within +/-2.7mm error, enabling automatic temperature measurement for mass screening of people at public places.

**Acknowledgment**

The authors would like to thank Dr. Rama Chellappa, Chair of the Department of Electrical and Computer Engineering, University of Maryland for his guidance and support during the research work.

### REFERENCES

1. IEC/ISO, "IEC 80601-2-59, Medical electrical equipment - Part 2-59: Particular requirements for the basic safety and essential performance of screening thermographs for human febrile temperature screening," (International Electrotechnical Commission, International Organization for Standardization, 2008).

2. B. Zitová, and J. Flusser, "Image registration methods: a survey," Image and Vision Computing **21**, 977–1000 (2003).
3. J. P. W. Pluim, J. B. A. Maintz, and M. A. Viergever, "Mutual-information-based registration of medical images: a survey," IEEE Transactions on Medical Imaging **22**, 986-1004 (2003).
4. R. P. Woods, J. C. Mazziotta, and S. R. Cherry, "MRI-PET registration with automated algorithm," Journal of Computer Assisted Tomography **17**, 536–546 (1993).
5. R. P. Woods, S. R. Cherry, and J. C. Mazziotta, "Rapid automated algorithm for aligning and reslicing PET images," Journal of Computer Assisted Tomography **16**, 620–633 (1992).
6. D. L. Hill, C. R. Maurer Jr., A. J. Martin, S. Sabanathan, W. A. Hall, D. J. Hawkes, D. Rueckert, and C. L. Truwit, "Assessment of intraoperative brain deformation using interventional MR imaging," in *International Conference on Medical Image Computing and Computer-Assisted Intervention*(Springer Berlin Heidelberg, 1999), pp. 910-919.
7. D. Mattes, D. R. Haynor, H. Vesselle, T. Lewellen, and W. Eubank, "Non-rigid multimodality image registration," in *Medical Imaging 2001*(International Society for Optics and Photonics, 2001), pp. 1609–1620.
8. J. Nocedal, and S. Wright, *Numerical optimization* (Springer Science & Business Media, 2006).
9. J.-P. Thirion, "Image matching as a diffusion process: an analogy with Maxwell's demons," Medical Image Analysis **2**, 243-260 (1998).
10. X. Gu, H. Pan, Y. Liang, R. Castillo, D. Yang, D. Choi, E. Castillo, A. Majumdar, T. Guerrero, and S. B. Jiang, "Implementation and evaluation of various demons deformable image registration algorithms on a GPU.," Phys Med Biol **55**, 207 (2009).
11. B. K. Horn, and B. G. Schunck, "Determining optical flow," Artificial intelligence **17**, 185-203 (1981).
12. J. Ma, J. Zhao, Y. Ma, and J. Tian, "Non-rigid visible and infrared face registration via regularized Gaussian fields criterion," Pattern Recognition **48**, 772-784 (2015).